\documentclass[conference]{IEEEtran}
\IEEEoverridecommandlockouts
\usepackage{cite}
\usepackage{amsmath,amssymb,amsfonts}
\usepackage{algorithmic}
\usepackage{graphicx}
\usepackage{textcomp}
\usepackage{xcolor}

\usepackage{multirow}
\usepackage{adjustbox}
\usepackage{tabularx}
\usepackage{booktabs}
\usepackage{enumitem}

\def\BibTeX{{\rm B\kern-.05em{\sc i\kern-.025em b}\kern-.08em
    T\kern-.1667em\lower.7ex\hbox{E}\kern-.125emX}}
\begin{document}

\title{Cross-resolution Face Recognition via Identity-Preserving Network and Knowledge Distillation\\
\thanks{Support from the Swiss National Science
Foundation (SNSF) 20CH21\_195532 for XAIface
CHIST-ERA-19-XAI-011 is acknowledged.}
}

\author{\IEEEauthorblockN{Yuhang Lu, Touradj Ebrahimi}
\IEEEauthorblockA{Multimedia Signal Processing Group (MMSPG)\\
\'Ecole Polytechnique F\'ed\'erale de Lausanne (EPFL)
}}

\maketitle

\begin{abstract}
Cross-resolution face recognition has become a challenging problem for modern deep face recognition systems. It aims at matching a low-resolution probe image with high-resolution gallery images registered in a database. Existing methods mainly leverage prior information from high-resolution images by either reconstructing facial details with super-resolution techniques or learning a unified feature space. 
To address this challenge, this paper proposes a new approach that enforces the network to focus on the discriminative information stored in the low-frequency components of a low-resolution image. 
A cross-resolution knowledge distillation paradigm is first employed as the learning framework. Then, an identity-preserving network, WaveResNet, and a wavelet similarity loss are designed to capture low-frequency details and boost performance. Finally, an image degradation model is conceived to simulate more realistic low-resolution training data. Consequently, extensive experimental results show that the proposed method consistently outperforms the baseline model and other state-of-the-art methods across a variety of image resolutions.

\end{abstract}


\section{Introduction}
\label{sec:intro}
In the past decades, face recognition (FR) has founds its way in many everyday applications. Current state-of-the-art deep learning-based face recognition systems achieve near-perfect performance on well-known public face recognition benchmarks such as LFW \cite{huang2008labeled} and MegaFace \cite{kemelmacher2016megaface}.
However, these face datasets are primarily collected in controlled environments and in high resolution, which quite differ from face images captured by real-world devices. 
In fact, studies \cite{zou2011very, cheng2019low, grm2018strengths,knoche2021image,lu2022novel} have demonstrated a significant performance deterioration of the most advanced deep face recognition systems in presence of resolution discrepancies. 
In this work, we mainly focus on the problem of cross-resolution face recognition (CRFR), which intends to match low-resolution (LR) probe images with high-resolution (HR) gallery images in a database. 

Most existing approaches to cope with CRFR can be divided into two categories. 
In the first category, HR images are reconstructed from LR images with face super-resolution (FSR) techniques \cite{zhang2018super, hsu2019sigan, kong2019cross, lai2019low, yin2020fan}, which are then recognized by a face recognition model. Although FSR methods can generate missing information, even facial details, they are mainly optimized for visual appearance and often ignore and even alter crucial identity information. This results in limited improvement of performance in LR domains. Furthermore, the high computational cost of the FSR module during both training and inference lays an additional burden on the entire face recognition system and heavily impairs its efficiency.

Different from FSR-based approaches, the second category converts LR and corresponding HR faces into a unified resolution-invariant feature space. These approaches rely fully on the identity information and learn a discriminative representation. Earlier work \cite{yang2017discriminative} leveraged a multidimensional scaling approach to learn a mapping matrix. Lu et al. \cite{lu2018deep} proposed a deep coupled ResNet model with two additional branch networks to map coupled HR and LR features to a common space.
Zangeneh et al. \cite{zangeneh2020low} directly employed a two-branch structure DCNNs to learn a non-linear feature transformation. 
\cite{kim2021quality} conceived an invertible decoder and learned a quality-agnostic model. 
\cite{zha2019tcn,lai2021deep,knoche2022octuplet} tackled the problem with a metric learning approach. They were all built on triplet loss and learned to reduce the resolution gap by pulling together positive HR-LR pairs and pushing away dissimilar ones. 

\begin{figure*}[t]
	\centering
	\begin{adjustbox}{width=0.95\textwidth}
    \includegraphics[]{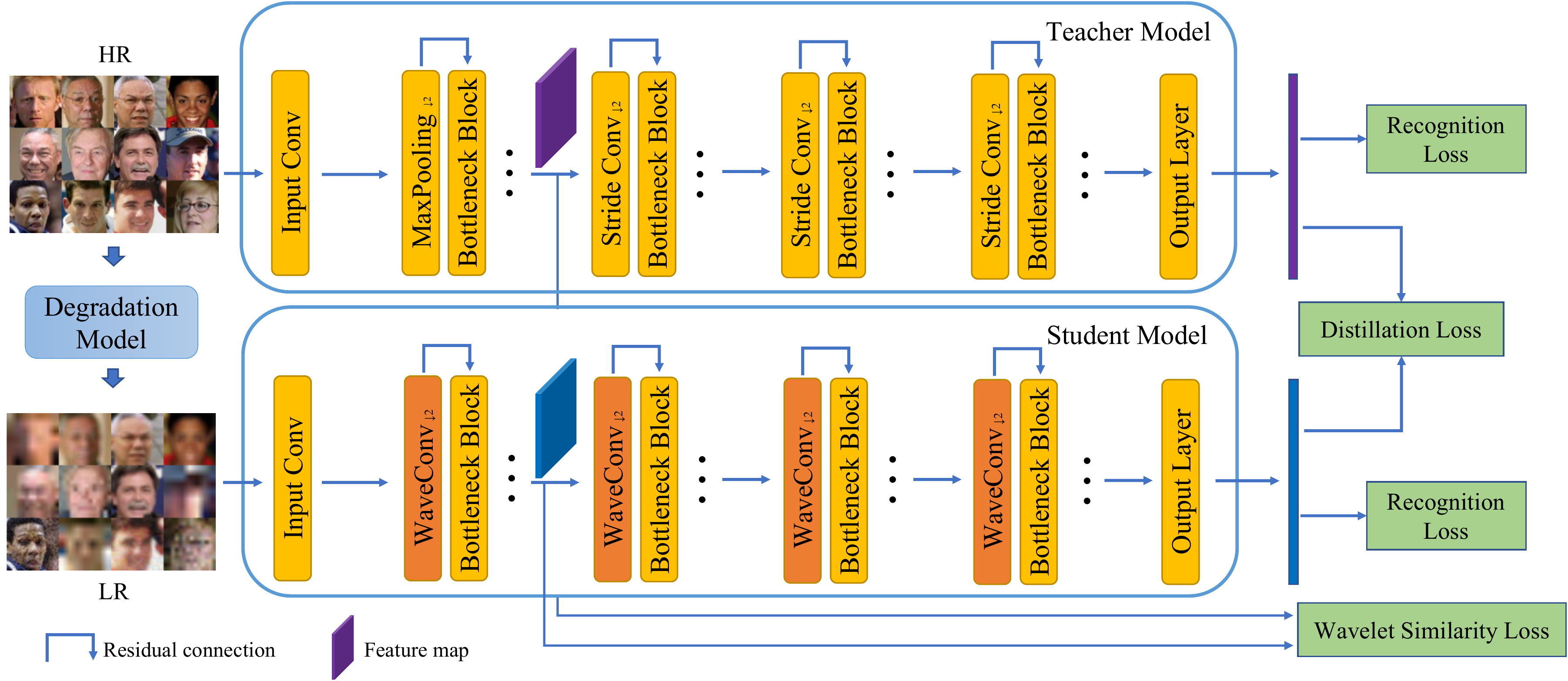}
	\end{adjustbox}
	\caption{The architecture of the proposed knowledge distillation framework and the identity-preserving student network.}
	\label{fig:kdframework}
\end{figure*}

Knowledge distillation is a typical approach that builds resolution-invariant feature space by distilling HR domain knowledge to the LR domain. This idea was first proposed in \cite{44873} to transfer knowledge from a high-performing but computationally expensive teacher network into a simpler student network. 
Recent studies \cite{zhu2019low,ge2018low,ge2020look, massoli2020cross,feng2021resolution,shin2022teaching,huang2022feature} have shown the potential of this approach in solving recognition problems in low-resolution domains. For instance, Zhu et al. \cite{zhu2019low} and Huang et al. \cite{huang2022feature} addressed the low-resolution object recognition problem with the teacher-student learning paradigm. 
Authors in \cite{ge2018low, wang2019improved,feng2021resolution} developed efficient low-resolution face recognition models at low computational cost by distilling informative facial features from teacher to a lightweight student network. 
Ge et al. \cite{ge2020look} obtained better performance in CRFR by distilling structural relationships across teacher and student networks. 
More recently, \cite{shin2022teaching} performed an attention similarity knowledge distillation. Instead of the feature map, they transferred attention maps obtained from the teacher network into a student network to boost performance in the LR domain. 
In this paper, a cross-resolution knowledge distillation framework is first employed, where the targeted student network is trained with multi-scale LR data and optimized with both face recognition and distillation losses.

Despite the guidance of the prior knowledge extracted from HR face images, the large resolution disparity between HR and LR images makes it difficult for the student network to capture informative features. Frequency analysis in \cite{knoche2021image, li2020wavelet} has shown that the high-frequency information in a facial image, such as edge and noise, is eliminated during the resolution reduction, while the low-frequency subbands still preserve the most discriminative features. 
Thus, this work proposes an identity-preserving network, WaveResNet, to capture the discriminative information stored in low-frequency components of the LR images. It is adapted from ResNet \cite{he2016deep} by replacing the pooling and stride convolution layers with a low-pass filter based on Discrete Wavelet Transform (DWT). 
The high-frequency subbands of the intermediate feature maps are filtered out to remove ambiguous and noisy information and enforce the network to focus on the more discriminative low-frequency information. 
In addition, a wavelet similarity loss is designed as an auxiliary distillation loss in order to further enhance attention in low-frequency subbands. 
Moreover, a degradation model is designed to simulate real-world LR training data and develop a more robust recognition system. The proposed method has been evaluated on four datasets in a variety of resolutions and it outperforms the baseline model and some other related solutions.


\section{Method}
\label{sec:method}

\subsection{Problem Definition}
This paper mainly describes and resolves the cross-resolution face recognition (CRFR) problem, where the probe images are LR due to the limited definition of the camera or the large distance between the camera and the subject, while the gallery images registered in the database are all of higher quality and resolution. 
In the testing phase, one focuses on the face verification task and examines the matching between an LR probe image and an HR gallery image. 

\subsection{Identity Preserving Network}

Different from FSR-based methods which aim at recovering high-frequency details for identity matching, this paper proposes to focus on the information stored in the low-frequency domain and directly mines deep identity features from LR training data. The main insight is that the high-frequency details are eliminated after the resolution reduction while the low-frequency components in LR images contain more discriminative information. 
In this subsection, an identity-preserving network, WaveResNet, is introduced for this purpose. The idea is to remove the ambiguous and noisy high-frequency information and enhance the discriminative features in LR images during the training process. Ideally, it performs more accurate recognition across various image resolutions. 



In detail, as shown in Fig.~\ref{fig:kdframework}, a low-pass convolutional filter based on Discrete Wavelet Transformation (DWT) is embedded into ResNet, denoted as WaveConv. It replaces the Maxpooling and stride convolution operations. Given an input image $\boldsymbol{x}$, a low-pass filter $\boldsymbol{f_{LL}}$ based on 2D DWT converts $\boldsymbol{x}$ into its low-frequency subband image $\boldsymbol{x_{LL}}$. The filter itself is a stride 2 convolutional operator during the transformation and automatically downsamples the image by a factor 2. The embedded operation in the WaveConv layer is defined as $\boldsymbol{x_{LL}}=(\boldsymbol{f_{LL}} \circledast \boldsymbol{x})\downarrow_{2}$, where $\circledast$ refers to convolution operator and $\downarrow_{2}$ means downsampling by 2. 

\subsection{Knowledge Distillation Framework for CRFR}

\subsubsection{Face Recognition Framework}
Fig. \ref{fig:kdframework} illustrates the knowledge distillation framework for the CRFR task. The teacher model is built on the ResNet network. Different from many teacher-student frameworks where the student model is a much simpler network for the sake of efficiency, our student network utilizes the proposed WaveResNet with the same amount of parameters to pursue high representation capability in both HR and LR data. Under this framework, the teacher network is first trained on HR images and learns to extract rich and informative facial details from high-quality training data. Then, cross-resolution distillation adapts the knowledge of discriminative features to the student network, which is trained on multi-scale LR data. 

\subsubsection{Losses} 
Under the framework of knowledge distillation, the following loss functions have been conceived and employed.

\noindent\textbf{Recognition Loss:}
The popular ArcFace \cite{deng2019arcface} loss is employed by both teacher and student networks as a recognition loss to learn the discriminative power.
\begin{align}
    \mathcal{L}_{arcface}=-\frac{1}{N} \sum_{i=1}^N \log \frac{e^{s\left(\cos \left(\theta_{y_i}+m\right)\right)}}{e^{s\left(\cos \left(\theta_{y_i}+m\right)\right)+\sum_{j=1, j \neq y_i}^n e^{s\left(\cos \left(\theta_j\right)\right)}}}.
\end{align}

\noindent\textbf{Cross-resolution Distillation Loss:}
In the training stage, the knowledge from the teacher network is transferred to the student model with a distillation loss. 
In order to improve the performance of the student network on different sizes of LR data, the distillation process is designed in a way to enforce a constraint over features across variant resolutions in one unified feature space. During the training, multi-scale versions of LR training data is collected and a cross-resolution distillation loss is applied to minimize the discrepancy between HR and LR features.
Specifically, given a pair of training samples, HR image $\boldsymbol{x_H}$ and LR image $\boldsymbol{x_L}$ of random size $s$, they are respectively passed into the teacher and student networks including classification layers to obtain the logits $\boldsymbol{z_H}$ and $\boldsymbol{z^s_L}$ and to calculate the loss. 
The distillation loss is expressed as:
\begin{align}
    \mathcal{L}_{distill} = \frac{1}{N} \sum_{i=1}^N  T^2 \mathcal{L}_{K L}\left(\sigma\left(\frac{z_H}{T}\right), \sigma\left(\frac{z^s_L}{T}\right)\right),
\end{align}
where $\mathcal{L}_{K L}$ refers to the KL Divergence, T is the temperature parameter to smooth the distillation loss, and $\sigma(\cdot)$ refers to the softmax function.

\noindent\textbf{Wavelet Similarity Loss:}
An additional auxiliary loss on the intermediate feature maps is introduced to further enhance the attention on low-frequency components, namely wavelet similarity loss. It enforces the student network to learn more discriminative knowledge stored in low-frequency features from the teacher network.
First, the feature maps from both teacher and student streams are spotted and then decomposed into multiple frequency bands by DWT. Afterward, MSELoss is applied to the low-frequency components only. The formula of the proposed wavelet similarity loss is as follows.
\begin{align}
    \mathcal{L}_{wavesim} = \sum_{k=1}^2 \lVert f_{LL}(z_H^k) - f_{LL}(z_L^k)  \rVert_2^2,
\end{align}
where $z^k$ means the intermediate feature in the $k^{th}$ stage of ResNet and $f_{LL}$ refers to the DWT-based low-pass filter.

The total loss is a weighted combination of recognition loss, distillation loss, and wavelet similarity loss.
\begin{align}
\mathcal{L}_{total} = \mathcal{L}_{arcface}+\lambda_{1} \mathcal{L}_{distill} + \lambda_{2} \mathcal{L}_{wavesim}.
\end{align}

\begin{figure}[t]
	\centering
    \includegraphics[width=\linewidth]{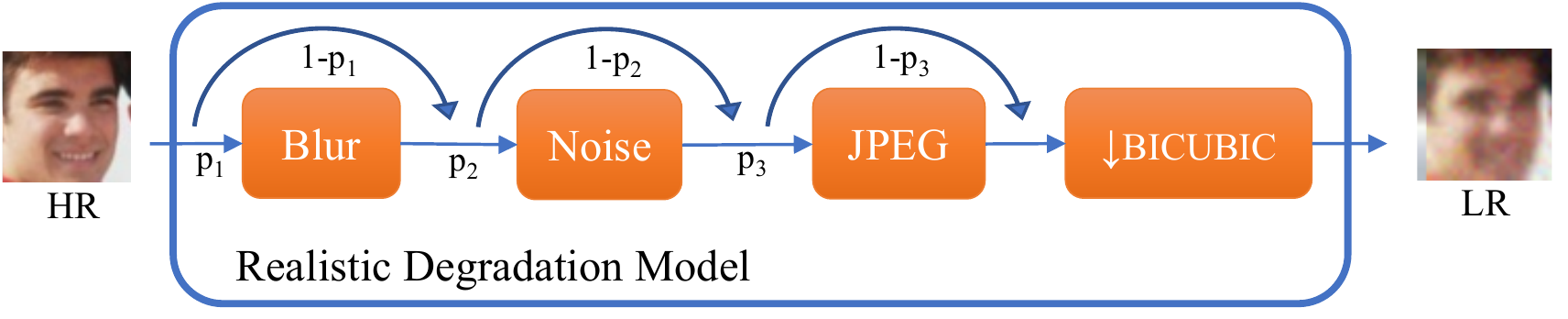}
	\caption{Data degradation model to produce realistic LR data.}
	\label{fig:degrade}
\end{figure}

\subsection{Degradation Model for LR Data Synthesis}
In the proposed learning framework, the student network is trained on synthesized low-resolution data. In order to develop a robust recognition system, the synthesized LR training data should not deviate much from those captured in the real world. 
Previous studies in the CRFR task tend to add Gaussian blur before downsampling to better simulate the low-resolution effect on images. In more realistic scenarios, LR images captured by surveillance cameras are often accompanied by random motion blur, noise, and compression artifacts. 
This paper hand-designs a degradation model to produce LR face images that are closer to real-world data. As depicted in Fig. \ref{fig:degrade}, the HR image is first randomly corrupted by blur operation, synthetic noise, and JPEG compression artifacts. During experiments, the probability of applying each corruption in the degradation model is set to 0.5. Afterward, the data is downsampled into selected sizes by bicubic operation. Some examples are visualized in Fig. \ref{fig:kdframework}.



\begin{table*}[t]
  \centering
  \caption{Verification accuracy (\%) of the proposed method on multiple datasets of different resolutions. Degradation means the degradation model. KD refers to the proposed knowledge distillation framework. WaveSim refers to the auxiliary wavelet similarity loss. 
  \textcolor[rgb]{ 1,  0,  0}{Red color} denotes the highest score and \textcolor[rgb]{ .267,  .447,  .769}{Blue color} denotes the second highest score. 
  }
    \begin{adjustbox}{width=0.85\textwidth}
    \begin{tabular}{cccc|cccc|c}
    \toprule
    \multicolumn{4}{c|}{Methods}          & \multicolumn{4}{c|}{Avg on $\cup$\{LFW,AgeDB,CPLFW,CALFW\}} & \multirow{2}[2]{*}{\shortstack{Overall\\ Average}} \\
    \cmidrule{1-8}
    Backbone  &  Degradation    & KD & WaveSim & \; 14x14 & 28x28 & 56x56 & 112x112\scriptsize(HR) &  \\
    \midrule
    ResNet       &       &       &       & 84.92 & \textcolor[rgb]{.267,.447,.769}{93.23} & 94.14 & 94.13 & 91.61 \\
    WaveResNet       &       &       &       & 86.73 & 92.86 & 94.17 & 94.17 & 91.98 \\
    WaveResNet       & \checkmark      &       &       & 87.77 & 92.55 & 93.31 & 93.30 & 91.73 \\
    WaveResNet       & \checkmark      & \checkmark      &       & \textcolor[rgb]{1,0,0}{88.31}  & 93.18  & \textcolor[rgb]{.267,.447,.769}{94.21} & \textcolor[rgb]{.267,.447,.769}{94.31} & \textcolor[rgb]{.267,.447,.769}{92.50} \\
    WaveResNet       & \checkmark      & \checkmark      & \checkmark      & \textcolor[rgb]{.267,.447,.769}{88.30} & \textcolor[rgb]{1,0,0}{93.25} & \textcolor[rgb]{1,0,0}{94.33} & \textcolor[rgb]{1,0,0}{94.47} & \textcolor[rgb]{1,0,0}{92.59} \\
    \bottomrule
    \end{tabular}%
    \end{adjustbox}
  \label{tab:mixdataset}%
\end{table*}%

\section{Experimental Results}
\label{sec:experiment}
\subsection{Experimental Settings}
\subsubsection{Datasets}
The cleaned MS1M dataset \cite{deng2019arcface} is used as the training set, which is composed of approximately 3.28M face images belonging to 72,778 identities. All the images in the training set are cropped to the size of 112x112 and aligned with five facial landmarks. Under the teacher-student training framework, every sample is randomly downsized in order to construct HR-LR training pairs. As for evaluation, four popular datasets are employed, i.e. LFW \cite{huang2008labeled}, AgeDB-30 \cite{moschoglou2017agedb}, CPLFW \cite{zheng2018cross}, and CALFW \cite{zheng2017cross}.
For a fair comparison with previous related work, all the testing samples are downsampled using linear interpolation instead of the degradation model. 

\subsubsection{Implementation Details}
In the proposed knowledge distillation framework, the teacher network leverages ResNet as a backbone and the student network employs the proposed WaveResNet. The teacher network is trained on HR images only, while the student network is trained on multi-scale LR images. The LR images are obtained through the proposed degradation model in random scales and then upsampled to the same size as HR images for training.
Both teacher and student networks are trained for 18 epochs using the SGD optimizer with a batch size of 128. The learning rate is initially set to be 0.1 and divided by 10 at 10, 13, and 16 epochs. The weights in the loss function are set to be $\lambda_{1}=1$ and $\lambda_{2}=0.05$. 


\subsection{Performance on Multiple Datasets}
Table. \ref{tab:mixdataset} shows the verification accuracy of the proposed method on multiple datasets of different resolutions. The results demonstrate the effectiveness of each proposed module. Compared to the baseline model in the first row, the proposed identity-preserving WaveResNet significantly improves the performance in very low-resolution testing images. Training with realistic synthetic data further improves the performance in LR test data but it impairs recognition accuracy on HR images. The cross-resolution distillation framework not only remedies the performance sacrifice in HR images but also enhances the accuracy in LR conditions, thereby improving the overall scores. 
Finally, after additionally employing the auxiliary wavelet similarity loss,
the model demonstrates promising results and significantly outperforms the baseline model on both low and high-resolution images.



\begin{figure}[t]
	\centering
    \includegraphics[width=\linewidth]{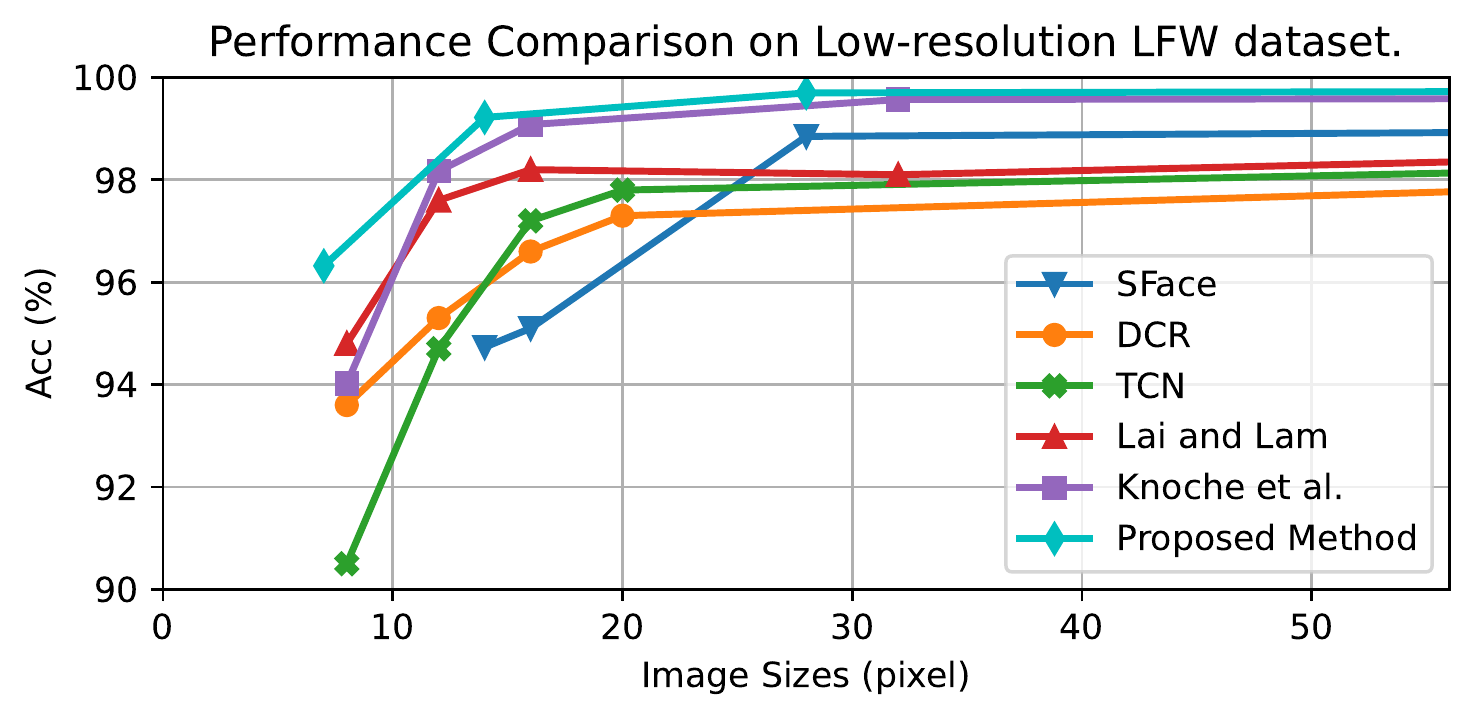}
	\caption{Verification accuracy (\%) on the LFW dataset.}
	\label{fig:lfw}
\end{figure}

\subsection{Comparison with the State-of-the-Art Methods}
The performance of the proposed method is also compared with other state-of-the-art approaches. Due to a lack of open-source codes, it is not possible to re-train the SOTA methods under exactly the same configurations. Thus, we directly took the highest-performing scores in their original publications for comparison. Fig. \ref{fig:lfw} presents the results of SFace\cite{lai2019low}, DCR \cite{lu2018deep}, TCN \cite{zha2019tcn}, Lai and Lam \cite{lai2021deep}, Knoche et al. \cite{knoche2022octuplet}, and our proposed method on low-resolution LFW dataset. The results show that our method consistently outperforms other approaches in both low and high-resolution settings. 
An additional comparisons with Kim et al. \cite{kim2021quality} and Shin et al. \cite{shin2022teaching} have been made on the AgeDB-30 dataset, see Table. \ref{tab:agedb}. As a result, the proposed method achieves the best performance across all resolutions of images. 
Besides, it is also observable that the FSR-based method \cite{lai2019low} performs better on higher-resolution data than many other approaches based on resolution-invariant feature spaces \cite{lu2018deep, zha2019tcn, lai2021deep}, but it is less powerful in very low-resolution scenarios. 





\begin{table}[t]
  \centering
  \caption{Verification accuracy (\%) on the AgeDB-30 dataset.}
    \begin{tabular}{ccccc}
    \toprule
    Methods & 14x14 & 28x28 & 56x56 & 112x112 \\
    \midrule
    Kim et al. \cite{kim2021quality} & 73.20 & 87.05 & 91.27 & 92.22 \\
    Shin et al.  \cite{shin2022teaching} & 79.45 & 89.15 & 93.58 & 93.78 \\
    Proposed Method  & 81.87 & 93.95 & 96.05 & 96.50 \\
    \bottomrule
    \end{tabular}%
  \label{tab:agedb}%
\end{table}%


\subsection{Discussion}
The experimental results demonstrate the effectiveness of the proposed method in the CRFR task. In fact, each module of the method plays a different role. 
For example, the WaveResNet and synthetic LR training data mainly contribute to LR face recognition, and the cross-resolution knowledge distillation paradigm further elevates the performance in HR images. The wavelet similarity loss additionally improves the performance on all resolutions. It is notable that most of the previous work presents a relatively poor result either in high or very low-resolution data. On the contrary, the proposed method offers high performance across a variety of resolution scenarios after combination of all proposed modules.


\section{Conclusion}
\label{sec:conclusion}


A new approach to address the challenge of cross-resolution face recognition was proposed based on identity-preserving network built upon a knowledge distillation framework. 
A realistic data degradation model is also contributed to further improve the performance in LR scenarios and demonstrating the discriminative power contained in the low-frequency domain of LR data. 







\bibliographystyle{IEEEtran}
\bibliography{refs}


\end{document}